\pdfoutput=1
\documentclass[11pt]{article}

\usepackage[]{acl}

\usepackage{times}
\usepackage{latexsym}

\usepackage[T1]{fontenc}
\usepackage[utf8]{inputenc}

\usepackage{microtype}

\usepackage{amsmath}% http://ctan.org/pkg/amsmath
\usepackage{amssymb,amsmath,amsthm,enumitem}
\usepackage{graphicx}
\usepackage{microtype}
\usepackage{longtable}
\usepackage{comment}
\usepackage{booktabs}
\usepackage{multirow}
\usepackage{rotating}

\def\equationautorefname~#1\null{Eq~#1\null}
\renewcommand{\sectionautorefname}{\S\kern-0.2em}
\renewcommand{\subsectionautorefname}{\S\kern-0.2em}
\renewcommand{\subsubsectionautorefname}{\S\kern-0.2em}

\usepackage{xcolor,colortbl}
\usepackage[absolute,overlay]{textpos}

\definecolor{darkergreen}{rgb}{0.0,0.4,0.0}
\definecolor{BurntOrange}{RGB}{255,112,52}

\usepackage{siunitx}
\usepackage{array}

\usepackage[tight,footnotesize]{subfigure}

\usepackage{url}

\title{What’s Different between Visual Question Answering\\for Machine ``Understanding'' Versus for Accessibility?}

\author{Yang Trista Cao\thanks{ $^{\star}$ Equal contribution}\\
  University of Maryland \\
  \texttt{ycao95@umd.edu} \\\And
  Kyle Seelman\footnotemark[1]\\
  University of Maryland \\
  \texttt{kseelman@umd.edu} \\\AND
  Kyungjun Lee\footnotemark[1]\\
  University of Maryland \\
  \texttt{kyungjun@umd.edu} \\\And
  Hal Daum\'e III \\
  University of Maryland \\
  Microsoft Research \\
  \texttt{me@hal3.name}\\}

\begin{document}
\maketitle
% \hfills\break

\begin{abstract}
In visual question answering (VQA), a machine must answer a question given an associated image. Recently, accessibility researchers have explored whether VQA can be deployed in a real-world setting where users with visual impairments learn about their environment by capturing their visual surroundings and asking questions. However, most of the existing benchmarking datasets for VQA focus on machine ``understanding'' and it remains unclear how progress on those datasets corresponds to improvements in this real-world use case. We aim to answer this question by evaluating discrepancies between machine ``understanding'' datasets (VQA-v2) and accessibility datasets (VizWiz) by evaluating a variety of VQA models. Based on our findings, we discuss opportunities and challenges in VQA for accessibility and suggest directions for future work.

\end{abstract}

\section{Introduction} \label{sec:tex/intro} % Motivated by the turing test (x), the field has been focusing on testing and improving bot's ability on image recognition, language understanding and generation, etc. 
% However, along with introduction of LNN, we have seen great improvement on bots' ability in accomplishing many goals and people started to apply these bots into real lifes. As ... we may want to shift our focus on building bots that can help people in real life situations.

\begin{figure*}[ht]
    \centering
    \includegraphics[width=0.95\textwidth,clip=true,trim=13 0 13 0]{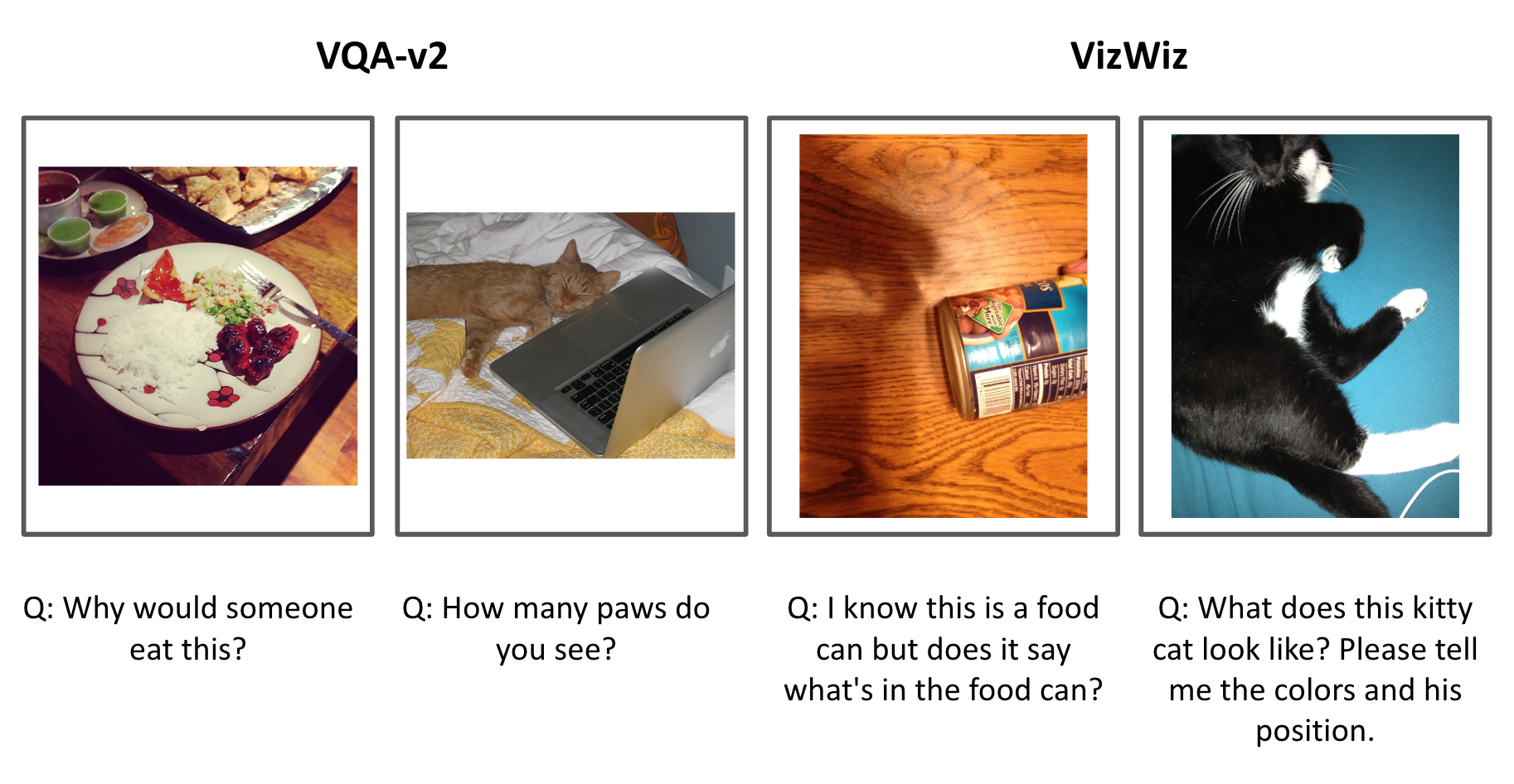} \vspace{-1em}
    \caption{Given similar image content (left: food, right: cat), questions in the machine ``understanding'' dataset VQA-v2 and the accessibility dataset VizWiz are substantially different. The VizWiz examples show questions that are significantly more specific (with one question even explicitly stating that it's already obvious that this is a can of food), more verbal, and significantly less artificial (as in the cat examples) than the VQA-v2 ones.}
    \label{fig:cat_food_example}
\end{figure*}

Much research has focused on evaluating and pushing the boundary of machine ``understanding'' -- can machines achieve high scores on tasks thought to require human-like comprehension, including image tagging and captioning \citep[e.g.,][]{lin2014microsoft}, and various forms of reasoning \citep[e.g.,][]{wang-etal-2018-glue,sap-etal-2020-commonsense}. 
In recent years, with the advancement of deep learning, we saw great improvements in machines' capabilities in accomplishing these tasks, raising the possibility for deployment.
However, adapting machine systems in real-life is non-trivial as real-life situations and users can be significantly different from synthetic and crowdsourced dataset examples \citep{Shneiderman_2020}.
In this paper we use the visual question answering (VQA) task as an example to call more attention to shifting from development on machine ``understanding'' to building machines that can make positive impacts to the society and people. 

Visual question answering (VQA) is a task that requires a model to answer natural language questions based on images. 
This idea dates back to at least to the 1960s in the form of answering questions about pictorial inputs~\citep[i.a.]{coles1968line,theune2007questions}, and builds on ``intelligence'' tests like the total Turing test~\citep{harnad1990symbol}.
Over the past few years, the task was re-popularized with new modeling techniques and datasets \citep[e.g.][]{daquar,okvqa}.
However, besides the purpose of testing a models' multi-modal ``understanding,'' VQA systems could be potentially beneficial for visually impaired people in answering their questions about the visual world in real-time. 
For simplicity, we call the former view \textit{machine understanding VQA} (henceforth omitting the scare quotes) and the latter \textit{accessibility VQA}.
The majority of research in VQA (\autoref{sec:tex/relate}) focuses on the machine understanding view.
As a result, it is not clear whether VQA model architectures developed and evaluated on machine understanding datasets can be easily adapted to the accessibility setting, as the distribution of images, questions, and answers might be---and, as shown in \autoref{fig:cat_food_example}, are---quite different. 
% \citet{vizwiz} introduced a VQA dataset that originated from visually-impaired people named VizWiz.

In this work, we aim to investigate the gap between the machine understanding VQA and the accessibility VQA by uncovering the challenges of adapting machine understanding VQA model architectures on an accessibility VQA dataset. 
Here, we focus on English VQA systems and datasets; for machine understanding VQA, we use the VQA-v2 dataset~\cite{vqa}, while for accessibility VQA, we use the VizWiz dataset~\cite{vizwiz} (\autoref{sec:datasets}).
% We use the VQA-v2 dataset~\cite{vqa} for machine understanding VQA, and the VizWiz dataset~\cite{vizwiz} is used for accessibility VQA.
% looking into the sighted VQA models' performance on the VizWiz dataset and 
%We first answer the question of how well do machine understanding VQA model architectures perform on the VizWiz dataset, and whether model advancements in machine understanding VQA systems are helpful for accessibility VQA (\autoref{sec:model_progress}). 
% \kyle{this is a grammar thing (which im terrible at) should these be question mark statements?} 
Through performance assessments of seven machine understanding VQA model architectures that span 2017--2021 (\autoref{sec:models}), we find that model architecture advancements on machine understanding VQA also improve the performance on the accessibility task, but that the gap of the model performance between the two is still significant and is \textit{increasing} (\autoref{sec:model_progress}). 
% This gap in accuracy on the two datasets indicates that the differences between the machine understanding and accessibility datasets are non-trivial \kyle{I think the "two tasks" is confusing. We state visual question answering is a "task" but that's it, so saying two I think is confusing. While I think I undertand it to mean, Vqa-v2 and vizwiz, I don't think that's clear and is not needed imo} \trista{I tried...}, and that adapting model architectures that were developed for machine understanding to assist visually-impaired people is challenging. 
This increasing gap in accuracy indicates that adapting model architectures that were developed for machine understanding to assist visually impaired people is challenging, and that model development in this area may indicate architectural overfitting.

% \trista{how about this? Cause we never mentioned the machine understanding dataset, I feel bit weird to say two datasets.}\kyle{i think this is fine}
We then further investigate what types of questions in the accessibility dataset remain hard for the state-of-the-art (SOTA) VQA model architecture (\autoref{sec:error_analysis}).
We adopt the data challenge taxonomies from \citet{bhattacharya2019does} and \citet{Zeng_2020} to perform both quantitative and qualitative error analysis based on these challenge classes.
We find some particularly challenging classes within the accessibility dataset for the VQA models as a direction for future work to improve on.
Additionally, we observe that many of the questions on which state-of-the-art models perform poorly are not due to the model not learning, but rather due to a need for higher quality annotations and evaluation metrics. 

\section{Related Work} \label{sec:tex/relate} To the best of our knowledge, this is the first work that attempts to quantify and understand the gap in performance VQA models have between the VQA-v2 dataset collected by sighted people and the VizWiz dataset that contains images and questions from people with visual impairments and answers from sighted people.  \citet{bradychi2013} conduct a thorough study on the types of questions people with visual impairments would like answered, and provide a taxonomy for the types of questions asked and the features of such questions. This work was a significant step in understanding the need in people with visual impairments for VQA systems. In combination with our own work, this gives a more complete picture of what kinds of questions not only contribute to better model performance, but actually help individuals with visual impairments. Additionally, \citet{Zeng_2020} seek to understand the task of answering questions about images from people with visual impairments (i.e., VizWiz) and those from sighted people (i.e., VQA-v2).
The authors identified the common vision skills needed for both scenarios and quantified the difficulty of these skills for both humans and computers on both datasets.

% datasets
\citet{vizwiz}, who published a very first visual question answering (VQA) dataset, ``VizWiz'' containing images and questions from people with visual impairments, pointed out the artificial setting of other VQA datasets that include questions that are artificially created by sighted people.
The VizWiz challenge is based on real-world data and directs researchers working on VQA problems toward real-world VQA problems. This dataset was built on data collected with a crowdsourcing app, where users with visual impairments share an image and a question with a sighted crowdworker who answers the question for them~\cite{bigham2010vizwiz}.
Other existing datasets, such as VQA~\cite{vqa_v1}, DAQUAR~\cite{daquar}, and OK-VQA~\cite{okvqa}, are different in that their questions were not provided by those who took images. Instead, the images were first extracted from web searches, and then questions were later provided by sighted crowdworkers who viewed and imagined questions to ask about those images.
Here, we see that people with visual impairments can benefit the most from VQA technology but most of the existing VQA datasets do not involve people with visual impairments.
%Comparing these datasets would allow researchers to understand differences and consequent effects on machine learning models for VQA applications. More specifically, we can evaluate a VQA model trained and tested on two different datasets, respectively---\textit{e.g.}, a VQA model can be trained on data from sighted crowdworkers and then tested on data from blind users, and vice versa.
%This analysis can help us view this problem from the perspective of parity of participation \cite{fraser2008abnormal} to understand how unbalanced participation affects such technology.

% what's done and what needs to be done
Some prior work has investigated VQA datasets further, focusing on assessing diversity in answers to visual questions. For instance, \citet{yang2018visual} looked at answers to visual questions created by blind people and sighted people and worked on anticipating the distribution of such answers. Predicting the distribution of answers asked, they helped crowdworkers create as many unique answers as possible for answer diversity.
\citet{bhattacharya2019does} tackle the same issue by looking at images of VQA. They proposed a taxonomy of nine reasons that cause differences in answers and developed a model predicting potential reasons that can lead to differences in answers. 
However, little work explores discrepancies between questions from actual users of VQA applications (i.e., users with visual impairments) and contributors who helped develop data for VQA applications.

Our work aims to understand this gap by assessing the discrepancies between the dataset containing artificially created data and the dataset containing real-world application data present across different VQA models. More specifically, we assess the performance of VQA models that were proposed in different times and delve into the old model and the state-of-the-art model with individual datapoints to identify patterns where the models perform poorly for the accessibility dataset.  
\section{Experiment Setup} \label{sec:tex/method} % \trista{I would make three subsections: 1. model 2. dataset and 3. Evaluation Metric. I like the dataset subsection. For the model subsection, maybe add a brief introduction to all the models, like this: \paragraph{MFH} \citep{} is a model ...}

% \begin{table*}[ht]
% \centering
% \begin{tabular}{c||c|c|c|c|c|c|c}
%                 & MFB   & MFH   & BAN   & BUTD  & MCAN  & Pythia & ALBEF (lg) \\
%     \hline
%     VQA-v2      & 65.2  & 66.2  & 66.0  & 63.7  & 67.6  & 66.9   & 71.6 \\
%     VQA-v2-sm   & 43.8  & 47.5  & 43.9  & 24.9  & 47.0  & 45.4   & 62.4 \\
%     VizWiz-ft   & 39.0  & 37.8  & 44.3  & 42.6  & 39.7  & 51.4   & 50.8   \\ 
%     VizWiz      & 43.2  & 44.8  & 44.9  & 44.7  & 48.6  & ---    & 48.8
%     % VizWiz-ft   & 31.1  & 31.7  & 34.7  & 33.8  & 32.2  & 51.4   & 50.8   \\ 
%     % VizWiz      & 33.2  & 34.3  & 34.2  & 34.7  & 36.6  & ---    & 48.8
% \end{tabular}
% \caption{\label{tab:model-performance} Model performances on the VQA-v2 dataset and the VizWiz dataset. The models are ordered by the time they were proposed. (Todo make a plot with it.)}
% \end{table*}

To evaluate how existing VQA models' performance on machine understanding dataset align with performances on the accessibility dataset, we select two VQA datasets and seven VQA models. 
One of the datasets, VQA-v2, was proposed for machine understanding, whereas the other dataset, VizWiz, was collected to improve accessibility for visually-impaired people. 
The seven VQA models, selected from the VQA-v2 leaderboard\footnote{\url{https://paperswithcode.com/sota/visual-question-answering-on-vqa-v2-test-dev}}, include MFB~\cite{yu2017mfb}, MFH~\cite{yu2018mfh}, BAN~\cite{kim2018ban}, BUTD~\cite{anderson2018bottom}, MCAN~\cite{yu2019mcan}, Pythia~\cite{jiang2018pythia}, and ALBEF~\cite{ALBEF}.
We assess all seven models on both of the datasets to investigate and understand the model progress across the machine understanding and accessibility datasets\footnote{Code is available at \url{https://github.com/kyleseelman/vqa_accessibility}}. 

\subsection{Datasets}
\label{sec:datasets}

As a representative of machine understanding VQA, we take the \textbf{VQA-v2} dataset~\cite{vqa}, which includes around 204,000 images from the COCO dataset~\cite{lin2014microsoft} with around one million questions. The images are collected through Flickr by amateur photographers. Thus the images are from sighted people rather than visually-impaired people. 
In addition, questions in VQA-v2 are collected in a post-hoc manner --- given a image, sighted crowdworkers are asked to create potential questions that could be asked for the image. 
Finally, given the image-question pairs, a new set of annotators are asked to answer the questions based on the image information. For each image-question pair, ten annotations are collected as ground-truth.

% Moreover, we form the other subset for further error analysis (Section~\ref{sec:error_analysis}). This subset is formed based on the annotations from prior work investigating factors that could lead to differences in VQA answers \cite{bhattacharya2019does} and annotating visual skills required to answer VQA questions \cite{Zeng_2020}. In our error analysis, datapoints with these annotations become a new test set, and the original train set without these datapoints is then defined as a new train set.
% \kyungjun{the paragraph above is added, please check.}
% \trista{I feel it may be more natural to talk about these in the experiment section?}
% \kyle{I think we can briefly mention them since they are datasets we used, but give the reasoning in the experiments section. Ex. stop after "...is the same as that of the VizWiz's train set" and before "In our error analysis..." }
% \kyungjun{Now we mention this small dataset again in the experiment section.}

As a representative of accessibility VQA, we take the \textbf{VizWiz} dataset~\cite{vizwiz}, which includes around 32,000 images and question pairs from people with visual impairments.
This dataset was built on data collected with a crowdsourcing-based app~\cite{bigham2010vizwiz} where users with visual impairments ask questions by uploading an image with a recording of the spoken question. 
The VizWiz dataset uses the image-question pairs from the data collected through the app and asks crowdworkers to annotate answers.
Similarly, ten ground-truth answers are provided for each image-question pair.
Note that in VizWiz each image-question pair is provided simultaneously by the same person, which is different from how the VQA-v2 dataset was curated. 

Our evaluation also uses a smaller subset of VQA-v2's training set, which we call \textit{VQA-v2-sm}, limited in size to match that of VizWiz's training set. This dataset is created to evaluate the effects of dataset size in VQA models' performance.

\subsection{Evaluation Metric}
% TODO: describe the experimental setup in detail, e.g.) answer space for VQA and VizWiz, and so on.
% We first evaluated the six different VQA models on two different VQA datasets, VQA-v2~\cite{vqa} and VizWiz~\cite{vizwiz}, in context of visual question answering. This evaluation employs the accuracy score that is widely used as the performance metric for benchmarking VQA models.%, which follows the equation shown in Eq.(TODO: do we need to present this formula?)
% For all the models, the answer space of the VQA-v2 dataset is $3,129$, while the answer space of the VizWiz dataset is $7,371$ (provided by Pythia~\cite{jiang2018pythia}). %TODO: need to check the answer space of the VQA-v2 dataset for Pythia.

We evaluate the seven models on the VQA-v2 and the VizWiz datasets with the standard ``accuracy'' evaluation metric for VQA. 
Since different annotators may provide different but valid answers, the metric does not penalize for the predicted answer not matching all the ground truth answers.
For each question, given the ten ground-truth from human annotators, we compute the model answer accuracy as in \autoref{eqa:accuracy_metric}. 
If the model accurately predicts an answer that matches at least three ground-truth answers, it receives a maximal score of $1.0$. Otherwise, the accuracy score is the number of ground-truth answers matched, divided by three:
\begin{equation}\label{eqa:accuracy_metric}
    \textrm{accuracy} = \min\left\{1, \frac{\textrm{\# matches}}{3}\right\}
\end{equation}

\begin{figure*}[ht]
    \centering
    \includegraphics[width=\textwidth,clip=true,trim=20 150 30 50]{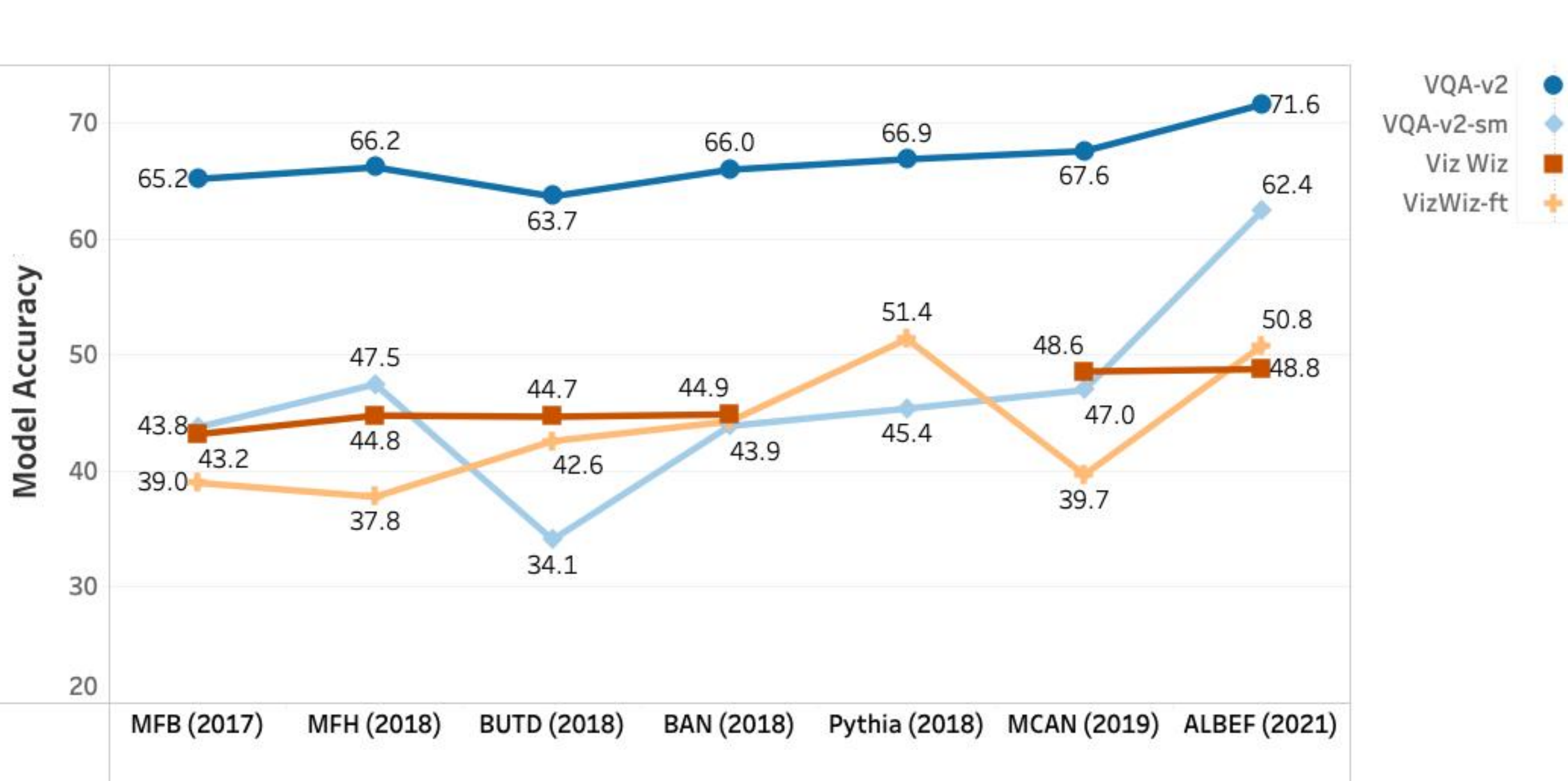}
    \caption{Model accuracy on VQA-v2 (including a sm(aller) subsampled version), and VizWiz (including a fine-tuned variant). The models are ordered by the time they were proposed. Improvements on VQA-v2 \emph{have} resulted in improvements on VizWiz, though the gap between the two remains significant.}
    \label{fig:model_perform}
\end{figure*}

\subsection{Models}
\label{sec:models}
All of the following models approach the problem as a classification task by aggregating possible answers from the training and validation dataset as the answer space. 
% \trista{This is my understanding. Can you verify if this is correct?} 
% \kyle{correct}
% \kyungjun{I think it also uses answers from the validation set, should we specify that as well?}
% \kyle{Fixed.}
%See appendix \autoref{sec:implementation} for additional implementation details.
% (TODO, also put the answer space size in appendix) --- \kyungjun{I mentioned the answer spaces in the Accuracy in VQA section, instead}
% For all the models, the answer space of the VQA-v2 dataset is $3,129$, while the answer space of the VizWiz dataset is $7,371$ (provided by Pythia~\cite{jiang2018pythia}). %TODO: need to check the answer space of the VQA-v2 dataset for Pythia.

\begin{description}[leftmargin=1em,labelwidth=0em,itemsep=0.25em,parsep=0em,topsep=0.5em,partopsep=0em]
\item[MFB \& MFH:] The multi-modal factorized bilinear \& multi-modal factorized high-order pooling models~\citep{yu2017mfb,yu2018mfh} are built upon the multi-modal factorized bilinear pooling that combines image features and text features as well as a co-attention module that jointly learns to generate attention maps from these multi-modal features. The MFB model is a simplified version of the MFH model.

\item[BUTD:] The bottom-up and top-down attention model~\cite{anderson2018bottom} goes beyond top-down attention mechanism and proposes the addition of a bottom-ups attention that finds image regions, each with an associated feature vector, thus, creating a bottom-up and top-down approach that can calculate at the level of objects and other salient image regions.

\item[BAN:] The bilinear attention network model~\cite{kim2018ban} utilizes bilinear attention distributions to represent given vision-language information seamlessly. BAN considers bilinear interactions among two groups of input channels, while low-rank bilinear pooling extracts the joint representations for each pair of channels. 

\item[Pythia:] Pythia is an extension of the BUTD model, utilizing both data augmentation and ensembling to significantly improve VQA performance~\cite{jiang2018pythia}.

\item[MCAN:] The modular co-attention network model~\cite{yu2019mcan} follows the co-attention approach of the previously mentioned models, but cascades modular co-attention layers at depth, to create an effective deep co-attention model where each MCA layer models the self-attention of questions and images.

\item[ALBEF:] The align before fusing model~\cite{ALBEF} builds upon existing methods that employ a transformer-based multimodal encoder to jointly model visual tokens and word tokens, by aligning the image and text representations and fusing them through cross-model attention. 
\end{description}

% MMNasNet~\cite{yu2020mmnasnet} is built with a deep multimodal neural architecture search algorithm called \textit{MMnas} powered by deep encoder-decoder model that finds predefined primitive operations for a target task (\textit{i.e.}, VQA). 

\noindent
For all the models, the answer space of the VQA-v2 dataset is $3,129$, while the answer space of the VizWiz dataset is $7,371$, which is provided by Pythia~\cite{jiang2018pythia}. 

\paragraph{Implementation details.} We use three different code bases for our evaluation: OpenVQA\footnote{\url{https://github.com/MILVLG/openvqa}}, Pythia\footnote{\url{https://github.com/allenai/pythia}}, and ALBEF\footnote{\url{https://github.com/salesforce/ALBEF}.}.
On the OpenVQA platform, four VQA models---MFB, BAN, BUTD, and MCAN---are already implemented. 
Pythia supports both of the VQA-v2 and Vizwiz datasets, but OpenVQA and ALBEF only support the VQA-v2 dataset. Thus, we implement the support of the VizWiz dataset on OpenVQA (i.e., for MFB, BAN, BUTD, and MCAN) and ALBEF.
Their default hyperparameters are used to train models on VQA-v2 and VizWiz, respectively.
For OpenVQA and ALBEF on which we implement the VizWiz support, the default hyperparameters for VQA-v2 are used to train models on VizWiz as well.
We fix the default accuracy metric implemented in OpenVQA, which is silently incompatible with the VizWiz data format, consistently underscoring predictions.

%Since different annotators may provide different but valid answers, the metric gives credit to model answers if it matches with at least three of the ground-truths provided by annotators.
% If an answer matches at least three of the annotators it is treated as a perfect score.

%In order to be consistent with ``human accuracies,' machine accuracies are averaged over all 10 choose 9 sets of human annotators.
% \trista{Please modify if this is incorrect.}
% \kyle{Modified to accurately reflect evaluation + some. Probably will come back to better rephrase this, essentially if it matches 3 then 100\%, otherwise it's a fractional of 1/3}

\section{Findings and Discussion} \label{sec:tex/finding}

\begin{figure*}[t]
    \centering
    \includegraphics[width=\textwidth,clip=true,trim=0 100 0 80]{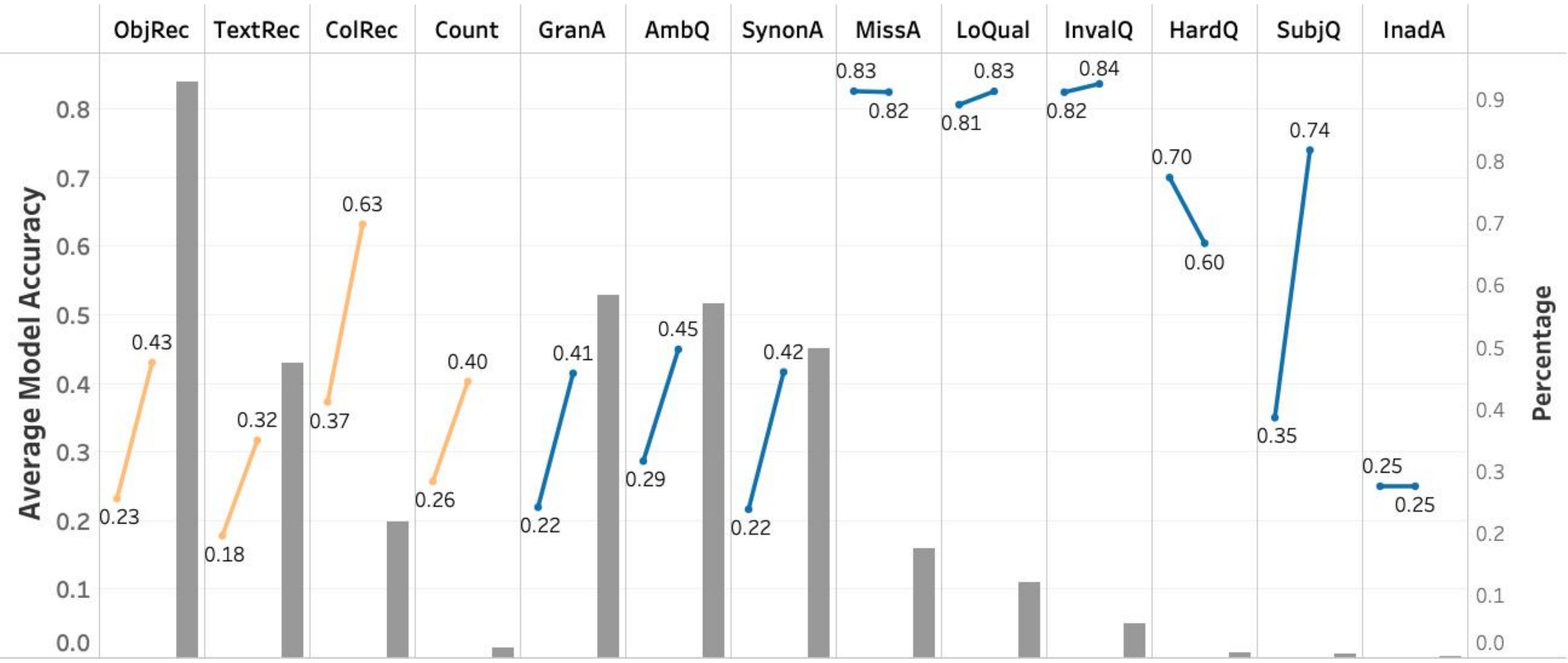}
    \caption{Model accuracy progress from MFB in 2017 (left point) to ALBEF in 2021 (right point) represented as lines measuring average model accuracy (left y-axis); these are subdivided by challenge classes from \citet{Zeng_2020} (the orange lines in \textit{ObjRec}--\textit{Count}) and from \citet{bhattacharya2019does} (the blue lines in \textit{GranA}--\textit{InadA}) for the VizWiz dataset. The bars represent the percentage of validation data examples that belong to the challenge classes (right y-axis). }
    \label{fig:progress_challenges}
\end{figure*}

Our objective in this section is to investigate challenges of the VQA task on two different datasets. We assess the performance progress of VQA models and delve into errors. Then, we discuss research directions that future work could take.

\subsection{Model Performance Progress}\label{sec:model_progress}
First, we examine whether the progress of VQA model architectures on the machine understanding dataset (VQA-v2) also apply to the accessibility dataset (VizWiz).
% We evaluate the seven VQA models introduced above on both the VQA-v2 and VizWiz datasets. 
For VizWiz, we report testing results on both trained from scratch with VizWiz (\textit{VizWiz}) and trained on VQA-v2 and finetuned with VizWiz (\textit{VizWiz-ft}).
As mentioned in Section~\ref{sec:datasets}, we randomly sampled the same number of datapoints from the train set of VQA-v2 as that in VizWiz to form \textit{VQA-v2-sm} to understand the effect of dataset size in the VQA performance. 
% \trista{I somehow feel this should go later when we use VQA-sm?}

The results are shown in \autoref{fig:model_perform}.\footnote{We do not include results of Pythia trained from scratch on VizWiz because their code expected to train VizWiz from a VQAv2.0 checkpoint, not from scratch. }
% Overall, the performance improvement is observed on the VQA-v2 dataset as well as the VizWiz dataset although the accuracy scores on the VizWiz dataset are much lower than those on the VQA-v2.
Overall, we observe that along with the advancement of model structures based on the VQA-v2 dataset, the model accuracy also improves on the VizWiz dataset.
% In particular, the model performance generally increases on data from sighted people (i.e., VQA-v2), and the state-of-the-art model (i.e., ALBEF) in this evaluation shows the highest performance on this dataset. However, the latest model does not perform best on data from people with visual impairments, compared to the other models. 
% From the oldest model (MFB) to the latest model (ALBEF), the performance increases by 6.4 points on VQA-v2 but only by 2.7 points on VizWiz---indeed, the accuracy score drops on the latest model by 0.7 points from MCAN and 15.5 points from Pythia.
% It indicates works that improve the VQA-v2 benchmark are beneficial in real-life application that assists visually-impaired people.  
% However, we also detect significant performance gaps between the models trained on the two datasets. The best model for VizWiz has accuracy score $51.4$ (from Pythia), which is still significantly lower than the MFB model's score on VQA-v2 in 2015 (with score gap of $14$). 
We observe that, from 2018 through 2021, performance on VQA-v2 improved $10\%$ relatively (from $65.2\%$ to $71.6\%$ accuracy), resulting in a similar improvement of $11\%$ ($43.8\%$ to $48.8\%$) on VizWiz without fine-tuning and 30\% ($39\%$ to $50.8\%$) on VizWiz with fine-tuning. The models fine-tuned from VQA-2 to VizWiz (i.e., VizWiz-ft) have similar performance with models trained on VizWiz from scratch. \citet{vizwiz} also reported a similar pattern but pointed out the gap between model performance and human performance.
These results show that improvements on VQA-v2 \textit{have} translated into improvements on VizWiz, whereas the performance gap between the two datasets are still significant.
% \hal{i feel like these sentences say the same thing multiple times. i would just say much more concrete things, like: from 2018 through 2021, performance on VQAv2 improved $10\%$ (from $65.2\%$ to $71.6\%$ accuracy), resulting in a similar improvement of $11\%$ ($43.8\%$ to $48.8\%$) on VizWiz without fine-tuning and 30\% ($39\%$ to $50.8\%$) on VizWiz with fine-tuning.
% These results show that overall improvements on VQAv2 \textit{have} translated into improvements on VizWiz.
% However, when controlling for dataset size, the we see an improvement of $42\%$ ($43.8\%$ to $62.4\%$) on VQA-v2-sm, where the training data is capped at the size of VizWiz, a substantially larger improvement than the $11\%$ seen on VizWiz (the result on VizWiz with fine-tuning is not comparable here, because it is fine-tuned from the full VQA-v2 dataset).
% This appears to demonstrate a significant ``overfitting'' effect, as both VQA-v2-sm and VizWiz start at almost exactly the same accuracy ($43.8\%$ and $43.2\%$) but performance on VQA-v2-sm improves significantly more than on VizWiz.
% }
% \hal{TODO check that the 43.8 is the right number for MFB on both VizWiz and VQA-v2-sm.}

% \hal{ah i should have kept reading before commenting, but perhaps somewhat can integrate what i wrote with what comes above and below.}

However, when controlling for dataset size, we see an relative improvement of $42\%$ ($43.8\%$ to $62.4\%$) on VQA-v2-sm, where the training data is capped at the size of VizWiz, a substantially larger improvement than the $11\%$ seen on VizWiz (the result on VizWiz with fine-tuning is not comparable here, because it is fine-tuned from the full VQA-v2 dataset).
This appears to demonstrate an ``overfitting'' effect, as both VQA-v2-sm and VizWiz start at almost exactly the same accuracy ($43.8\%$ and $43.2\%$) but performance on VQA-v2-sm improves significantly more than on VizWiz.

% A possible reason of the significant score gap is the training dataset size difference. 
% VizWiz dataset includes around $32,000$ image and question pairs whereas VQA-v2 includes around $1,000,000$ pairs. 
% Thus, to perform a fair comparison, we re-train the models on \textit{VQA-v2-sm}. %a random subset of VQA-v2 dataset to match the size of the VizWiz dataset.
% \autoref{fig:model_perform} includes the models performance on \textit{VQA-v2-sm}. 
% We see that models performance with VQA-v2-sm drop significantly to the level of model scores on VizWiz. 
% This indicates that training data size is a primary factor of the performance gap between VQA-v2 and VizWiz.
% Meanwhile, training models on the original VQA-v2 dataset and finetuned on the VizWiz dataset (i.e. VizWiz-ft) does not help to improve the model performance on the VizWiz dataset. 
% This demonstrates that simply adapting the machine understanding dataset (i.e. VQA-v2) for accessibility VQA is not enough. 
% We see that though the VQA-v2-sm performance is lower than the original VQA-v2 score, it is still significantly higher than the model performance on the VizWiz dataset. 
% Overall, this indicates that the two datasets have more salient discrepancies other than dataset size, which makes it non-trivial to adapt models from VQA-v2 to VizWiz dataset.

\subsection{Error Analysis}
\label{sec:error_analysis}
We perform both quantitative and qualitative error analysis to better understand which types of data will be useful to improve accessibility VQA for future dataset collection and model improvement. In this section we discuss the overall patterns found for models evaluated on VizWiz and what type of questions specifically, these model fail on.

\subsubsection{VQA Challenge Datasets}

In our first set of experiments, we aim to understand more precisely what that models have improved on between 2017 and 2021 that has led to an overall accuracy improvement on VizWiz-ft from $39.0\%$ (MFB) to $50.8\%$ (ALBEF).
To do this, we make use of two meta-data annotations of a subset of the VizWiz validation dataset ($3,143$ data examples): one labels each example with the \textit{vision skills} required to answer that question~\cite{Zeng_2020}, the second labels each with aspects of the \textit{image-question} combination that are challenging~\cite{bhattacharya2019does}. Both of these papers investigate the challenges for \textit{annotators}; here, we use these annotations to evaluate models.
\autoref{tab:error_label} shows the taxonomies of VizWiz validation examples that are labeled with the challenge class according to majority vote over five annotations. 

\begin{figure*}[t]
    \centering
    \includegraphics[width=\textwidth,clip=true,trim=0 170 0 150]{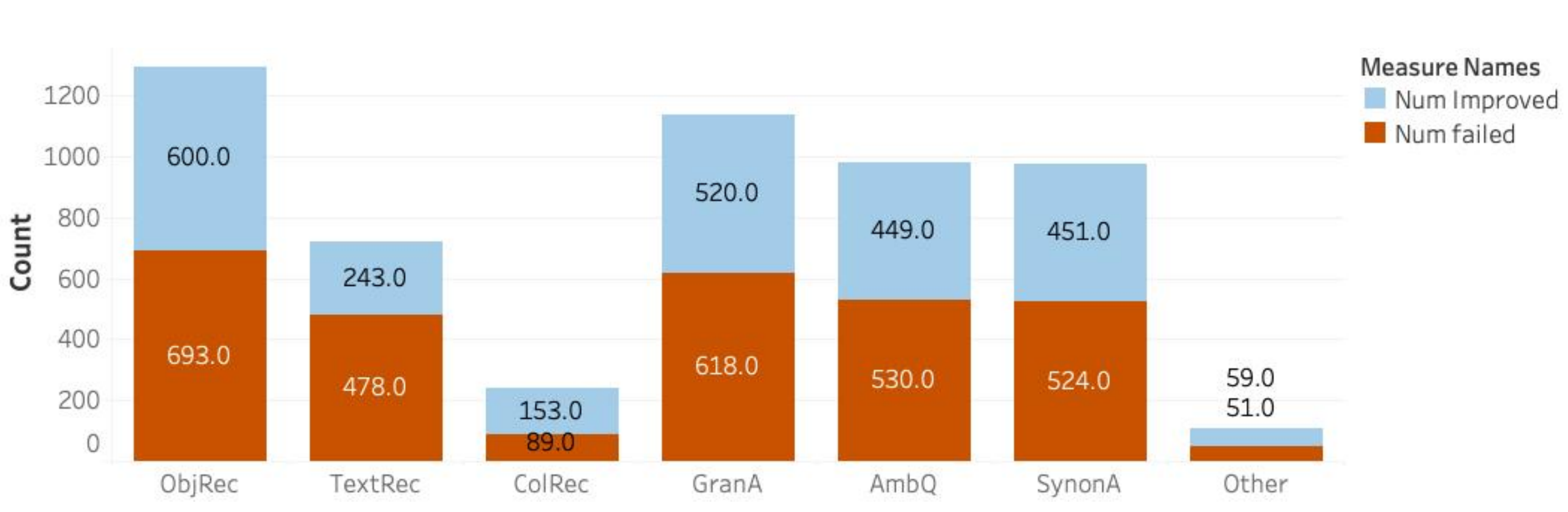}
    \caption{The performance improvements from MFB to ALBEF on the VizWiz dataset with respect to the reduced number of data examples with $0$ accuracy score. Red color represents the number of data examples ALBEF got $0$ score on while red plus blue color represents the number of data examples MFB got $0$ score on -- blue color thus represent the number of data examples improved by ALBEF from MFB. Note that we combine the challenge classes that has less than $50$ data examples as ``Other''. }
    \label{fig:exact_improve}
\end{figure*}

\begin{table}[t]
    \centering
    
    %ObjRec, ColorRec, TextRec, Count
    %LoQual, MissingA, InvQ, DiffQ, AmbigQ, SubjQ, SynonA, GranA, InadeqA
    \begin{tabular}{@{~}c@{~~~~}c@{~~~~}p{5.2cm}@{~}}
        \toprule
        & \textbf{Label} &  \textbf{Definition}  \\
        \midrule
        \multirow{4}*{\begin{turn}{90}\textbf{Vision}\end{turn}}
        &ObjRec &  object recognition  \\
        &TextRec &  text recognition \\
        &ColRec &  color recognition \\
        &Count &  counting \\
        \midrule
        \multirow{9}*{\begin{turn}{90}\textbf{Image-Question}\end{turn}}
        &GranA &  answers at different granularities \\
        &AmbQ &  ambiguous qs w/ $>1$ valid answer\\
        &SynonA &  different wordings of same answer\\
        &MissA &  answer not present given image\\
        &LoQual &  low quality image\\
        &InvalQ &  invalid question\\
        &HardQ &  hard question requiring expertise\\
        &SubjQ &  subjective question\\
        &InadA &  inadequate answers\\
        \bottomrule
    \end{tabular}
    \caption{VQA challenge taxonomies with labels. }
    \label{tab:error_label}
\end{table}

Given this taxonomy, we assess the performance progress between MFB and ALBEF in the VizWiz-ft setting across each VQA challenge class.
The results are reported in \autoref{fig:progress_challenges}.
Compared to MFB, ALBEF improves on every class of challenges except HardQ---hard questions that may require domain expertise, special skills, or too much effort to answer---though HardQ is also one of the rarest categories. (It is somewhat surprising the high performance of the models on these ``hard'' questions.)
We observe that among the \textit{vision skill} challenge classes, the models struggle the most on recognizing texts.
Among the \textit{image-question} challenges, models have low accuracy on almost all the challenge classes related to the answers --- ground-truth answers with different granularities, wordings, and inadequate answers. 
This indicates a potential problem in evaluating models on the VizWiz dataset, which is further explored in our qualitative analysis in \autoref{sec:where-fail}.
For the questions, models struggle the most with handling ambiguous or subjective questions, which we will discuss more in the next section. 
Overall, the results point out the challenges that models have most difficulty on, which we hope can bring insights for future work to improve accessibility VQA systems.

% We also look into data examples within each challenge class that the VizWiz-ft model get low score on. We randomly pick 10 examples from each challenge class. TODO are some examples. We find that todo.

\subsubsection{Where the Models Fail}\label{sec:where-fail}

\begin{figure*}[t]
    \centering
    \includegraphics[width=\textwidth,,clip=true,trim=5 5 5 5]{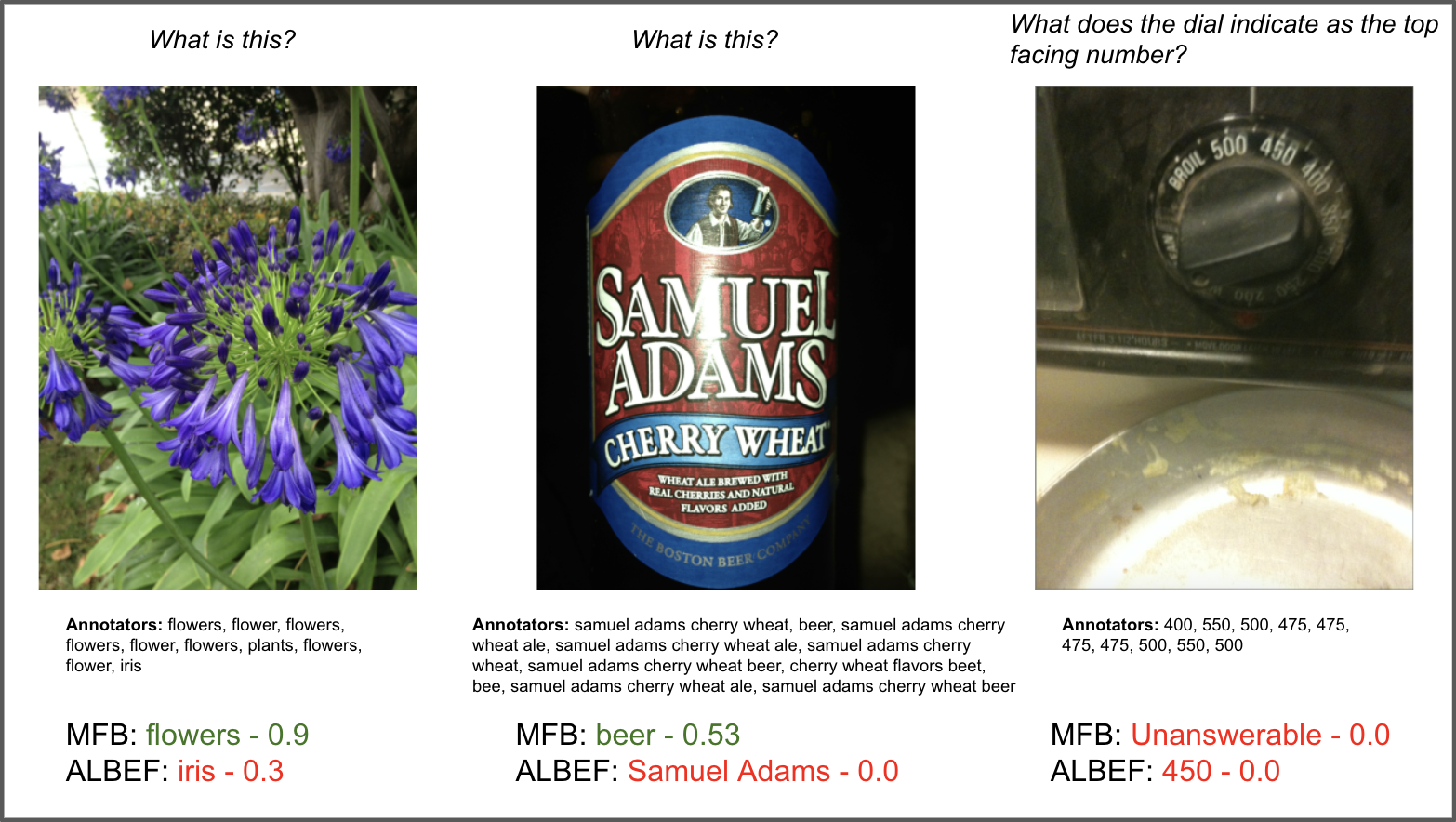}
    \caption{Examples of low-performance ALBEF image/question pairs, that should be correct, together with the accuracy scores. (Left) ALBEF gives a more detailed answer of \textit{Iris}, but since most annotators put \textit{flowers}, performance score is low. (Middle) ALBEF correctly names the beer, but once again does not match the annotators, so MFB appears to perform better. (Right) ALBEF gives a number answer that is close to correct (and which is not much different from the set of ground truth answers), where MFB does not make an attempt. }
    \label{fig:error_analysis}
\end{figure*}

To further understand the data examples that the models fine-tuned on VizWiz perform poorly on, we manually investigate the validation examples on which models achieve $0\%$ accuracy: matching none of the ten human-provided answers.
We measure how many data examples that have $0\%$ accuracy on MFB got improved by the ALBEF model for each challenge class, shown in  \autoref{fig:exact_improve}.

Model improvement is greatest on color recognition ($63\%$) and least on text recognition ($34\%$).  
Meanwhile, object recognition, text recognition, color recognition, and ambiguous questions are the challenge classes which a current state-of-the-art model has the most difficulty.
% \hal{plz replace COL, OBJ, etc. in all the figures with the labels from Table 1, and sort in the same order}
When taking a closer look at the individual examples that ALBEF has $0\%$ accuracy on, it turns out the issue is often with the \textit{evaluation measure} and not with the ALBEF model itself. The most frequent issues are:

\paragraph{Answerable Questions Marked Unanswerable.}
The biggest difference (and what we deem an improvement) between ALBEF and MFB has to do with ``unanswerable'' questions. $27\%$ of the questions in the validation data are deemed ``unanswerable'' by at least three annotators---making ``unanswerable'' a prediction that would achieve perfect accuracy.
For 56\% of the questions that were not of type \textit{``unanswerable''}, MFB still answered \textit{``unanswerable''}, while ALBEF did this only 30\% of the time. This skew helps MFB on the evaluation metric, but ALBEF's answers for many of these questions are at least as good---and therefore useful to a user---as saying ``unanswerable.'' For example, the \textit{number} question type, MFB only answered with a number $2.2\%$ of the time, whereas ALBEF answered with a number $56\%$ of the time and, in those cases, the answers are often very close to the correct answer (see \autoref{fig:error_analysis}). %This does not automatically mean ALBEF always does better, sometimes "unasnwerable" is still scored higher than the answer ALBEF chooses. However, in these cases when manually analyzing the image/question pair, we find that ALBEF's answers sometime are better answer than the annotators (Figure \ref{fig:error_analysis}). 

\paragraph{Overly Generic Ground Truth.}
It is often the case that ALBEF provides a correct answer that is simply more specific than that provided by the ground truth annotation. For example, a common question in VizWiz is \textit{``What is this?''}. When comparing ALBEF and MFB models, by accuracy alone, ALBEF outperforms MFB in $28.8\%$ of such cases, and MFB outperforms ALBEF in $12.6\%$. However, in the majority of these examples, ALBEF gives a correct, but more detailed response than the ground truth, thus earning it $0\%$ accuracy (for example see \autoref{fig:error_analysis}). So while, based on the annotation,  ALBEF is wrong, the model is actually correctly answering the question and performs worse than the MFB model only 2.6\% of the time. 
Furthermore, we found that both MFB and ALBEF models are both challenged by \textit{yes/no} question types, but that these questions were often subjective or ambiguous.

\paragraph{Annotator Disagreement.}
Questions such as \textit{``Is this cat cute?''} or \textit{``Are these bad for me?''} arguably make for poor questions when evaluating model performance: highly subjective yes/no questions often have annotations where at least three annotators state \textit{``yes''}, and at least three state \textit{``no''}. Therefore, per the evaluation metric, either answer achieves an accuracy of $1$. For example, for the question \textit{``Do these socks match?''} ALBEF had an accuracy score of $60\%$ for an answer of \textit{no} and MFB had an accuracy score of $83\%$ for an answer of \textit{yes}, even though either is arguably correct.

%\subsection{Discussion and Moving Forward}

\section{Limitations} \label{sec:tex/limitation} This work aims to understand the degree to which progress on machine ``understanding'' VQA has, and has the ability to, improve performance on the task of accessibility VQA. Our findings should be interpreted with several limitations in mind. First, while we analyzed many models across several years of VQA research, our analysis is limited to two datasets. Moreover, as discussed in \autoref{sec:where-fail}, the ``ground truth'' in these datasets, especially when combined with the standard evaluation metric, is not always reliable. Second, our analysis is limited to English, and may not generalize directly to other languages. 
Finally, blind and low-vision users are not a monolithic group, and the photos taken and questions asked in the VizWiz dataset are representative only of those who used the mobile app, likely a small, unrepresentative subset of the population. 

\section{Conclusion and Future Directions} \label{sec:tex/conclusion} In this paper, we have shown that, overall, performance improvements on machine ``understanding'' VQA \textit{have} translated into performance improvements on the real-world task of accessibility VQA.
However, we have also shown evidence that there may be a significant overfitting effect, where significant model improvements on machine ``understanding'' VQA translate only into modest improvements in accessibility VQA.
This suggests that if the research community continues to only hill-climb on challenge datasets like VQA-v2, we run the risk of ceasing to make any process on a pressing human-centered application of this technology, and, in the worst case, could degrade performance.

We have also shown that along with the overall model improvement, the accessibility VQA system have improved on almost all of the challenge classes though some challenges remain difficult.
In general, we observe the models struggle most on questions that require text recognition skill as well as ambiguous questions. 
Future work thus may wish to pay more attention on these questions in both data collection and model design. 

Finally, we have seen that we are likely reaching the limit of the usefulness of the standard VQA accuracy metric, and that more research is needed to develop automated evaluation protocols that are robust and accurately capture performance improvements. On top of this, VQA systems are reaching impressive levels of performance, suggesting that human evaluation of their performance in ecologically valid settings is becoming increasingly possible. As ecological validity would require conducting such an evaluation with blind or low-vision users, research is needed to ensure that such evaluation paradigms are conducted ethically and minimize potential harms to system users.

%Overall, this \hal{what is "this"} shows that it is non-trivial to adapt models that were developed focusing on ``machine understaning'' to real-life applications as real-life data can be very different from crowdsourced synthetic data. 
%which is reasonable as VizWiz samples data from real-life situations while VQA-v2 is a crowdsourced dataset. 
%Talk about our findings that can help future work in improving VizWiz dataset.
% This suggests that in order to really help the people with visual impairments, the field should work on the real data from visually-impaired people rather than chasing leaderboards.  

\section*{Acknowledgments}

This material is based upon work partially supported by the National Science Foundation under Grant No. \textit{2131508}. The authors are also grateful to all the reviewers who have provided helpful suggestions to improve this work, and to Hernisa Kacorri and Jordan Boyd-Graber who have provided pointers and suggestions.

%\bibliographystyle{acl_natbib}
\bibliography{vqa}

%\appendix
%\includesection{tex/appendix}{Appendix}

\end{document}